\documentclass[graybox]{svmult}

\usepackage{mathptmx}       
\usepackage{helvet}         
\usepackage{courier}        
\usepackage{type1cm}        
\usepackage{makeidx}         
\usepackage{multicol}        
\usepackage[bottom]{footmisc}

\usepackage{epsfig}
\usepackage{graphicx}
\usepackage{wrapfig}

\usepackage{amsmath}
\usepackage{amssymb}
\usepackage{textcomp}
\usepackage{upgreek}
\usepackage{bm}

\usepackage{url}
\usepackage{xspace}

\usepackage{cite}

\usepackage{multirow}
\usepackage{rotating}
\usepackage{booktabs}
\usepackage{afterpage}

\usepackage{algorithm2e}
\usepackage{alltt}
\usepackage{listings}

\usepackage{color}

\usepackage{relsize}
\usepackage{mdwlist}
\usepackage{comment}

\usepackage[font=small]{caption}
\usepackage[font=small]{subcaption}

\usepackage{multirow}
\usepackage{rotating}
\usepackage{booktabs}
\usepackage{tabularx}

\usepackage{enumitem}
\usepackage[olditem,oldenum]{paralist}

\usepackage[pagebackref=true,breaklinks=true,letterpaper=true,colorlinks,bookmarks=false,citecolor=red]{hyperref}

\usepackage{mysymbols}

\newenvironment{packed_enum}{
\begin{enumerate}
  \setlength{\itemsep}{1pt}
  \setlength{\parskip}{0pt}
  \setlength{\parsep}{0pt}
}
{\end{enumerate}}

\newenvironment{packed_item}{
\begin{itemize}
  \setlength{\itemsep}{1pt}
  \setlength{\parskip}{0pt}
  \setlength{\parsep}{0pt}
}{\end{itemize}}

\newlength{\sectionReduceTop}
\newlength{\sectionReduceBot}
\newlength{\subsectionReduceTop}
\newlength{\subsectionReduceBot}
\newlength{\abstractReduceTop}
\newlength{\abstractReduceBot}
\newlength{\captionReduceTop}
\newlength{\captionReduceBot}
\newlength{\subsubsectionReduceTop}
\newlength{\subsubsectionReduceBot}

\newlength{\horSkip}
\newlength{\verSkip}

\newlength{\figureHeight}
\usepackage{listings}
\usepackage{xcolor}
\usepackage{setspace}
\usepackage{verbatim}

\colorlet{punct}{red!60!black}
\definecolor{background}{HTML}{EEEEEE}
\definecolor{delim}{RGB}{20,105,176}
\colorlet{numb}{magenta!60!black}

\lstdefinelanguage{json}{
    basicstyle=\normalfont\ttfamily,
    numbers=left,
    numberstyle=\scriptsize,
    stepnumber=1,
    numbersep=8pt,
    showstringspaces=false,
    breaklines=true,
    frame=lines,
    backgroundcolor=\color{background},
    literate=
     *{0}{{{\color{numb}0}}}{1}
      {1}{{{\color{numb}1}}}{1}
      {2}{{{\color{numb}2}}}{1}
      {3}{{{\color{numb}3}}}{1}
      {4}{{{\color{numb}4}}}{1}
      {5}{{{\color{numb}5}}}{1}
      {6}{{{\color{numb}6}}}{1}
      {7}{{{\color{numb}7}}}{1}
      {8}{{{\color{numb}8}}}{1}
      {9}{{{\color{numb}9}}}{1}
      {:}{{{\color{punct}{:}}}}{1}
      {,}{{{\color{punct}{,}}}}{1}
      {\{}{{{\color{delim}{\{}}}}{1}
      {\}}{{{\color{delim}{\}}}}}{1}
      {[}{{{\color{delim}{[}}}}{1}
      {]}{{{\color{delim}{]}}}}{1},
}

\lstnewenvironment{code}[1][]%
{
   \noindent
   \minipage{\linewidth} 
   \vspace{0.5\baselineskip}
   \lstset{basicstyle=\ttfamily\footnotesize,frame=single,#1}}
{\endminipage}

\definecolor{Code}{rgb}{0,0,0}
\definecolor{Decorators}{rgb}{0.5,0.5,0.5}
\definecolor{Numbers}{rgb}{0.5,0,0}
\definecolor{MatchingBrackets}{rgb}{0.25,0.5,0.5}
\definecolor{Keywords}{rgb}{0,0,1}
\definecolor{self}{rgb}{0,0,0}
\definecolor{Strings}{rgb}{0,0.63,0}
\definecolor{Comments}{rgb}{0,0.63,1}
\definecolor{Backquotes}{rgb}{0,0,0}
\definecolor{Classname}{rgb}{0,0,0}
\definecolor{FunctionName}{rgb}{0,0,0}
\definecolor{Operators}{rgb}{0,0,0}
\definecolor{Background}{rgb}{0.98,0.98,0.98}

\lstnewenvironment{python}[1][]{
\lstset{
numbers=none,
numberstyle=\footnotesize,
numbersep=1em,
xleftmargin=1em,
framextopmargin=2em,
framexbottommargin=2em,
showspaces=false,
showtabs=false,
showstringspaces=false,
frame=l,
tabsize=4,
basicstyle=\ttfamily\small\setstretch{1},
backgroundcolor=\color{Background},
language=Python,
commentstyle=\color{Comments}\slshape,
stringstyle=\color{Strings},
morecomment=[s][\color{Strings}]{"""}{"""},
morecomment=[s][\color{Strings}]{'''}{'''},
morekeywords={import,from,class,def,for,while,if,is,in,elif,else,not,and,or,print,break,continue,return,True,False,None,access,as,,del,except,exec,finally,global,import,lambda,pass,print,raise,try,assert},
keywordstyle={\color{Keywords}\bfseries},
morekeywords={[2]@invariant},
keywordstyle={[2]\color{Decorators}\slshape},
emph={self},
emphstyle={\color{self}\slshape},
}}{}

\makeindex             

\begin{document}

\title*{CloudCV: Large Scale Distributed Computer Vision as a Cloud Service}

\author{Harsh Agrawal, Clint Solomon Mathialagan, Yash Goyal, Neelima Chavali, 
Prakriti Banik, Akrit Mohapatra, Ahmed Osman, Dhruv Batra}

\authorrunning{Agrawal, Mathialagan, Goyal, Chavali, Banik, Mohapatra, Osman, Batra} 

\institute{Harsh Agrawal \at Virginia Tech, \email{harsh92@vt.edu}
\and Clint Solomon Mathialagan \at Virginia Tech \email{mclint@vt.edu} 
\and Yash Goyal \at Virginia Tech \email{ygoyal@vt.edu} 
\and Neelima Chavali \at Virginia Tech \email{gneelima@vt.edu} 
\and Prakriti Banik \at Virginia Tech \email{prakriti@vt.edu} 
\and Akrit Mohapatra \at Virginia Tech \email{akrit@vt.edu}
\and Ahmed Osman \at Imperial College London \email {ahmed.osman99@gmail.com} 
\and Dhruv Batra \at Virginia Tech \email{dbatra@vt.edu}}

%
%
\maketitle

\abstract*{We are witnessing a proliferation of massive visual data. 
Unfortunately scaling existing computer vision algorithms to large datasets leaves 
researchers repeatedly solving the same algorithmic, logistical, and infrastructural problems. 
Our goal is to democratize computer vision; one should not have to be a computer vision, 
big data and distributed computing expert to have access to state-of-the-art distributed computer 
vision algorithms. 
We present CloudCV, a comprehensive system to provide access to 
state-of-the-art distributed computer vision algorithms as a 
cloud service through a Web Interface and APIs.}

\abstract{We are witnessing a proliferation of massive visual data. 
Unfortunately scaling existing computer vision algorithms to large datasets leaves 
researchers repeatedly solving the same algorithmic, logistical, and infrastructural problems. 
Our goal is to democratize computer vision; one should not have to be a computer vision, 
big data and distributed computing expert to have access to state-of-the-art distributed computer 
vision algorithms. 
We present CloudCV, a comprehensive system to provide access to 
state-of-the-art distributed computer vision algorithms as a 
cloud service through a Web Interface and APIs.}

\section{Introduction}
\label{sec:intro}

A recent World Economic Form report~\cite{bigdata_wef12} 
and a New York Times article~\cite{bigdata_nyt12} 
declared data to be a new class of economic asset, like currency or gold. 
Visual content is arguably the fastest growing data on the web. 
Photo-sharing websites like Flickr and Facebook now host more than 6 and 90 Billion photos (respectively). 
Every day, users share 200 million more images on Facebook. 
Every minute, users upload 72 hours or 3 days worth of video to Youtube. Besides consumer data, diverse 
scientific communities (Civil \& Aerospace Engineering, Computational Biology, Bioinformatics, and 
Astrophysics, \etc) are also beginning to generate massive archives of visual content~\cite{pacs,bisque,Berriman_queue11}, 
without necessarily having access to the expertise, infrastructure and tools to analyze them. 

This data revolution presents both an opportunity and a challenge. 
Extracting value from this asset will require converting meaningless data into perceptual understanding and 
knowledge. This is challenging but has the potential to \emph{fundamentally change the way we live} 
-- from self-driving cars bringing mobility to the visually impaired, 
to in-home robots caring for the elderly and physically impaired, 
to augmented reality with Google-Glass-like wearable computing units.

\subsection{Challenges}

In order to convert this raw visual data into knowledge and intelligence, we need to address a 
number of key challenges:

\begin{packed_item}
\item \textbf{Scalability.}
The key challenge for image analysis algorithms in the world of big-data is scalability. 
In order to fully exploit the latest hardware trends, we must address the challenge of 
developing fully distributed computer vision algorithms. Unfortunately, scaling 
existing computer vision algorithms to large datasets leaves researchers repeatedly solving the 
same infrastructural problems: building \& maintaining a cluster of machines, 
designing multi-threaded primitives for each algorithm and distributing jobs, pre-computing \& caching 
features \etc.

Consider for instance the recent state-of-the-art image categorization system by the Google/Stanford
team~\cite{le_icml12}. The system achieved an impressive 70\% relative improvement over
the previous best known algorithm for the task of recognizing 20,000 object categories in the 
Imagenet dataset~\cite{imagenet}. To achieve this feat, the system required a sophisticated 
engineering effort in exploiting model parallelism and had to be trained on a cluster with 
2,000 machines (32,000 cores) for one week. While this is a commendable effort, 
lack of such an infrastructural support and intimate familiarity with parallelism 
in computer vision algorithms leaves most research groups marginalized, computer vision 
experts and non-experts alike. 

\item \textbf{Provably Correct Parallel/Distributed Implementations.}
Designing and implementing efficient and provably correct parallel computer vision algorithms 
is extremely challenging. Some tasks like extracting statistics from image collections are 
\emph{embarrassingly parallel}, \ie can be parallelized simply by distributing the images to different 
machines. This is where framework such as MapReduce have demonstrated success. 
Unfortunately, most tasks in computer vision and machine learning such as training 
a face detector are not embarrassingly parallel -- 
there are data and computational dependencies between images and various steps in the 
algorithm. Moreover, for each such 
parallel algorithm, researchers must repeatedly solve the same low-level problems: 
formulating parallelizable components 
in computer vision algorithms, designing multi-threaded primitives, 
writing custom hardware wrappers, implementing mechanisms to avoid race-conditions, dead-locks, \etc. 

\item \textbf{Reusability.}
Computer vision researchers have developed vision algorithms that solve specific 
tasks but software developers building end-to-end system find it extremely difficult to integrate 
these algorithms into the system due to different software stacks, dependencies and different 
data format. Additionally, hardware designers have developed various dedicated computer 
vision processing platforms to overcome the problem of intensive computation. However, 
these solutions have created another problem: heterogeneous hardware platforms have made 
it time-consuming and difficult to port computer vision systems from one hardware platform to another.


\end{packed_item}

\subsection{CloudCV: Overview}

\begin{figure*}[t]
\centering
\includegraphics[width=1\linewidth]{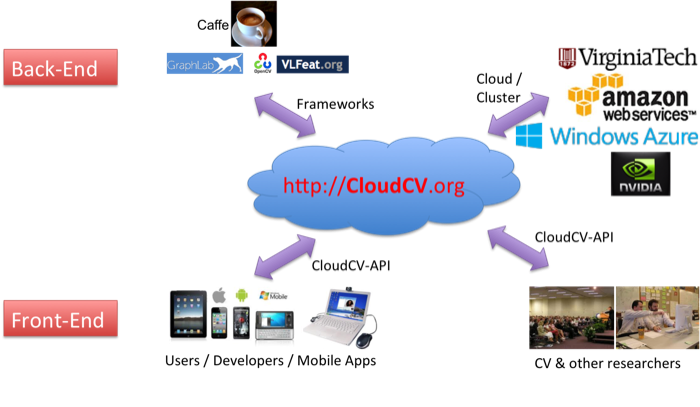}
\caption{Overview of CloudCV.}
\label{fig:overview}
\end{figure*}

In order to overcome these challenges, we are building 
\textbf{CloudCV}, a comprehensive system that will provide access 
to state-of-the-art distributed computer vision algorithms on the cloud. 


CloudCV today consists of a group of virtual machines running on Amazon Web Services capable of 
running large number of tasks in a distributed and parallel setting. Popular datasets used are already 
cached on these servers to facilitate researchers trying to run popular computer vision algorithms on 
these datasets. Users can access these services through a web interface which allows user to upload 
a few images from either Dropbox or local system and obtain results real-time. For larger datasets, 
the system enables to embed CloudCV services into a bigger end-to-end system by utilizing Python 
and Matlab APIs. Since the APIs are fairly easy to install through standard package managers, 
the researchers can now quickly run image analysis algorithms on huge datasets in a distributed 
fashion without worrying about infrastructure, efficiency, algorithms and technical know-how. 
At the back-end, on recieving the list of images and the algorithm that needs to be executed, 
the server distributes these jobs to worker nodes that process the data in parallel and communicate 
the results to the user in real time. Therefore, the user does not need to wait for the processing 
to finish on the entire dataset and can monitor the progress of the job due to real-time updates. 

\subsection{Application}

CloudCV will benefit three different audiences in different ways: 
\begin{packed_item}
\item \textbf{Computer vision researchers}: who do not have resources to or do not want to
reinvent a large-scale distributed computer vision system. For such users, CloudCV can  
serve as a unified data and code repository, providing cached version of all 
relevant data-representations and features. We envision a system where a
program running on CloudCV simply ``calls'' for a feature; if it is cached, 
the features are immediately loaded from distributed storage ( HDFS~\cite{hdfs}); if it is not cached, then 
the feature extraction code is run seamlessly in the background and the results are cached for future use. 
Eventually, CloudCV becomes the ultimate repository for ``standing on the shoulders of giants''.

\item \textbf{Scientists who are not computer vision experts:} but have large image collections 
that  needs to be analyzed. Consider a biologist who needs to automate the process of cell-counting
in microscopy images. Today such researchers must find computer vision 
collaborators and then invest in the logistical infrastructure required to run 
large-scale experiments. CloudCV can eliminate both these constraints, by 
providing access to state-of-the-art computer vision algorithms \emph{and}
 compute-time  on the cloud.
 
\item \textbf{Non-scientists:} who simply want to learn about computer vision
by demonstration. There is a tremendous demand from industry professionals and developers 
for learning about computer vision. 
Massive Open Online Classes (MOOCs) like Udacity and Coursera have demonstrated 
success. CloudCV can build on this success by being an important teaching tool for learning 
computer vision by building simple apps on the cloud. Imagine a student writing 4 lines of code in CloudCV
development environment to run a face-detector on a stream of images captured from his laptop webcam.

\end{packed_item}

\section{Related Efforts}
\label{related_work}

Before describing the architecture and capabilities of CloudCV, 
let us first put it in context of related efforts in this direction. 
Open-source computer vision software can be broadly categorized into three types:
\begin{itemize}

\item{General-Purpose Libraries:} There are a number of general purpose computer vision libraries 
available in different programming languages: 
\begin{packed_item}
\item{C/C++:} OpenCV~\cite{opencv}, IVT~\cite{ivt}, VXL~\cite{vxl}
\item{Python:} OpenCV (via wrappers), PyVision~\cite{pyvision}
\item{.NET:} AForge.NET~\cite{aforge}.
\end{packed_item}

The most comprehensive effort among these is OpenCV, which is a library aimed at real-time 
computer vision. It contains more than 2500 algorithms and has been downloaded 5 million
by 47K people~\cite{opencv}. The library has C, C++, Java and Python interfaces and runs on 
Windows, GNU/Linux, Mac, iOS \& Android operating systems.

\item{Narrow-Focus Libraries:} A number of toolboxes provide specialized implementations for 
specific tasks, \eg Camera Calibration Toolbox~\cite{calib3d}, 
Structure-from-Motion toolboxes~\cite{bundler, cmvs, visualsfm}, Visual Features Library~\cite{vlfeat}, 
and  deep learning frameworks such as Caffe~\cite{jia2014caffe}, Theano~\cite{Bastien-Theano-2012, bergstra+al:2010-scipy}, Torch~\cite{torch} \etc.

\item{Specific Algorithm Implementations:} released by authors on their respective websites. 
Popular examples include object detection~\cite{voc-release4}, articulated body-pose 
estimation~\cite{yang_cvpr11}, graph-cuts for image segmentation~\cite{Kolmogorov2004}, \etc. 

\end{itemize}

Unfortunately, all three source-code distribution mechanisms suffer from at least one of these limitations:
\begin{enumerate}

\item \textbf{Lack of Parallelism:} Most of the existing libraries have a fairly limited or no support for
parallelism. OpenCV and VLFeat for instance have multi-threading support, allowing programs to 
utilize multiple cores on a single machine. Unfortunately, modern 
datasets are so large that no single machine may be able to hold all the data. This makes it necessary to 
distribute jobs (with computational \& data dependencies) on a cluster of machines. CloudCV will have full
support for three levels of parallelism: i) single machine with multiple cores; ii) multiple machines in a cluster
with distributed storage; and iii) ``cloudbursting'' or dynamic allocation of computing resources via a professional elastic cloud computing service (Amazon EC2~\cite{ec2}). 

\item \textbf{Burden on the \emph{User} not the Provider:} Today, computer vision libraries place 
infrastructural responsiblities squarely on the user of these systems, not the provider. The user 
must download the said library, resolve dependencies, compile code, arrange for computing 
resources, parse bugs and faulty outputs. CloudCV will release the user of such 
burdens -- the user uploads the data (or points to a cached database in the 
repository) and simply specifies what computation needs to be performed. 

\end{enumerate}
Finally, we that stress CloudCV is not another computer vision toolbox. Our focus is not on
re-implementing algorithms, rather we build on the success of comprehensive efforts of OpenCV, Caffe
\& others. Our core contribution will be to provide fully distributed implementations on the 
cloud and make them available as a service.

\runinhead{Efforts Closest to the Goal of CloudCV:} There are multiple online services which provide specific algorithms such as face, concept, celebrity ~\cite{rekognition} or provide audio and video understanding~\cite{clarifai}, personalized object detectors~\cite{visionai}. 
Unlike these services, CloudCV is an open-source architecture that aims to provide 
the capability to run a user's own version of CloudCV on cloud services such as 
Amazon Web Services, Microsoft Azure, \etc. 

\section{CloudCV Back-end Infrastructure}

In this section, we describe in detail all the components that form the back-end architecture of CloudCV. 

\begin{figure*}[t]
	\centering
	\includegraphics[width=1\columnwidth]{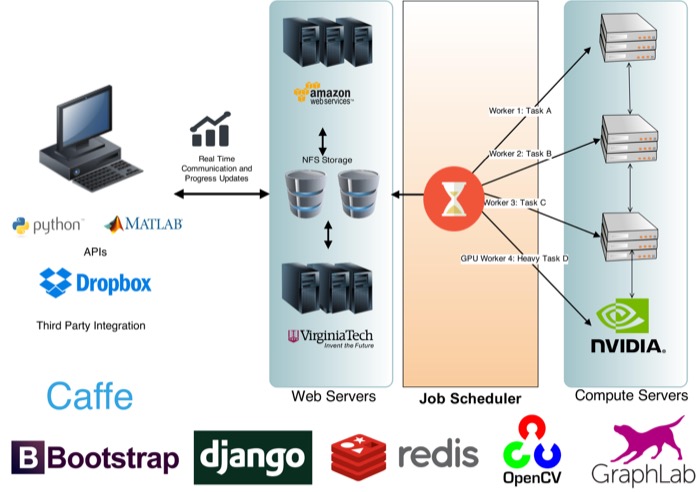}
	\caption{Users can access CloudCV using a web interface, Python or MATLAB API. The back-end consists of web servers which communicates with the client in real-time through HTTP and Web-sockets. The job schedule at the master node distributes incoming jobs across multiple computer servers (worker nodes). 
	\label{fig:backendflowchart}}
\end{figure*}

The back-end system shown in \figref{fig:backendflowchart} mainly consists of a web-server that is responsible for listening to 
incoming job requests and sending real-time updates to the user. 
A job scheduler takes these incoming jobs and distributes them across number of worker nodes. 
The system uses a number of open-source  frameworks to ensure an efficient design that can scale 
to a production system. 

\subsection{Web Servers}

The back-end consists of two servers that are constantly listening for incoming requests. 
We use a python based web-framework which handles Hypertext Transfer Protocol (HTTP) 
requests made by the web-interface or the APIs. These requests contains details about the 
job such as list of images, which executable to run, executable parameters, user information \etc. 
One drawback to HTTP requests is that it allows only a single request-response pair, \ie, for a 
given request the server can only return one response after which the connection breaks and 
the server cannot communicate with the client unless client sends a request. This leads to serious 
limitations because a persistent real-time connection cannot be established for the server to send 
updates to the user. To solve this problem, we use the web-socket protocol (Socket.IO) 
on top of another server (Node.js) 

\subsubsection{Django}

CloudCV uses Django~\cite{django} which is a high-level Python HTTP Web framework that is based on 
the Model View Controller (MVC) pattern. MVC defines a way of developing software so that 
the code for defining and accessing data (the model) is separate from request routing logic (the controller), 
which in turn is separate from the user interface (the view). 

A key advantage of such an approach 
is that components are loosely coupled and serve single key purpose. 
The components can be changed independently without affecting the other pieces. 
For example, a developer can change the URL for a given part of the application 
without affecting the underlying implementation. A designer can change a page'™s HTML code without 
having to touch the Python code that renders it. A database administrator can rename a database 
table and specify the change in a single place, rather than having to search and replace through a dozen files. 

Scaling up to serve thousands of web request is a crucial requirement. 
Django adopts a ``share nothing'' philosophy in which each part of the web-stack is broken 
down into single components so that inexpensive servers can be added or removed with minimum fuss. 

In the overall CloudCV back-end architecture, 
Django is responsible for serving the web-pages, translating requests into jobs and calling the 
job scheduler to process these jobs. The output of the algorithm that the job is running is pipelined 
to a message queue system. The receiver of the message queue system sends the output back to the user. 
In CloudCV, the message queue system is Redis and the receiver is Node.js; 
both of these are explained in the next two sections. 

\subsubsection{Node.js}

Node.js\cite{nodejs} is an event-driven web-framework that excels in real-time applications such as push updates 
and chat applications. 

CloudCV uses Node.js for real-time communication with the user so that all 
updates related to a particular job can be communicated to the user. 

Unlike traditional frameworks that use the stateless request-response paradigm such as Django, 
Node.js can establish a two-way communication with the client so that server can send 
updates to the client without the need for the client to query the server to check for updates. 
This is in contrast to the typical web response paradigm, where the client always initiates communication. 
Real time communication with the client is important because completing a job that contains large 
amounts of data will take some time and delaying communication with the client until the end of job is 
makes for a poor user experience and having the client query the server periodically is wasteful. 

The de facto standard for building real-time applications Node.js applications is via Socket.IO~\cite{socketio}. 
It is an event-based bi-directional communication layer which abstracts many low-level details 
and transports, including AJAX long-polling and WebSockets, into a single cross-browser 
compatible API. Whenever an event is triggered inside Node.js server, an event callback 
mechanism can send a response to the client. 

\subsubsection{Redis}


One of the use-cases of real time communication with the user is the 
ability to send algorithm output to the user during execution. 
To make this possible, there needs to be a system in place that can pipeline 
the algorithm output to the Node.js server, which is responsible for communicating the output back to client. 

In case of CloudCV, this system is Redis~\cite{redis}, a high-performance in-memory key-value data store. 
Since the data is stored in RAM (in-memory), looking up keys and returning a value is very fast. 

Redis can also act as a message queue between two processes -- worker process executing a 
particular algorithm and the Node.js server. Whenever a text output is generated by the executable, 
the worker process sends the output string through Redis. 
Node.js triggers an event whenever it receives the message from the message queue. 
The message consists of the output string, and the socket id of the client to which this output needs to be sent. Consequently, the event handler sends the output string to the user associated with that particular socket id. 

\begin{figure*}[t]
	\centering
	\includegraphics[width=1\textwidth]{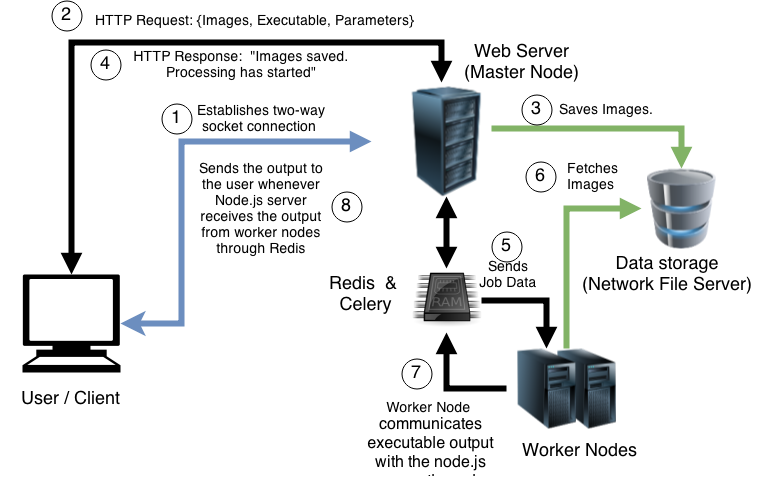}
	\caption{Flow describing the execution of a job starting from the user connecting to the CloudCV system to the back-end sending results to the user during execution of the job in real-time. 
	\label{fig:cloudcvBackendFlow}}
\end{figure*}

\figref{fig:cloudcvBackendFlow} describes the process of executing a job in detail. The flow is as follows:
\begin{enumerate}

\item At the start of a job, the user establishes a two-way socket connection with the server. 
Each user is recognized by the unique socket id associated with this connection. 

\item The details about the job such as list of images to process, name of the functionality 
that needs to be executed and its associated parameters are sent to the server using HTTP request. 

\item The server saves the image in the database and sends a response back to the user. 

\item The server then distributes the job to worker nodes by serializing all the data. 
An idle worker node pops the job from the queue, fetches the image from the network file server 
and starts executing the functionality associated with the job. 

\item Whenever the executable generates an output, the worker node informs the 
master node by sending the generated output through a message queue. 

\item Upon receiving the message, the master node sends the output to the client. 
This is made possible by the event-driven framework of Node.js (as explained in previous sections). 
\end{enumerate}

\subsection{Distributed Processing and Job Scheduler}

\subsubsection{Celery}

\begin{figure*}[t]
	\centering
	\includegraphics[width=1\columnwidth]{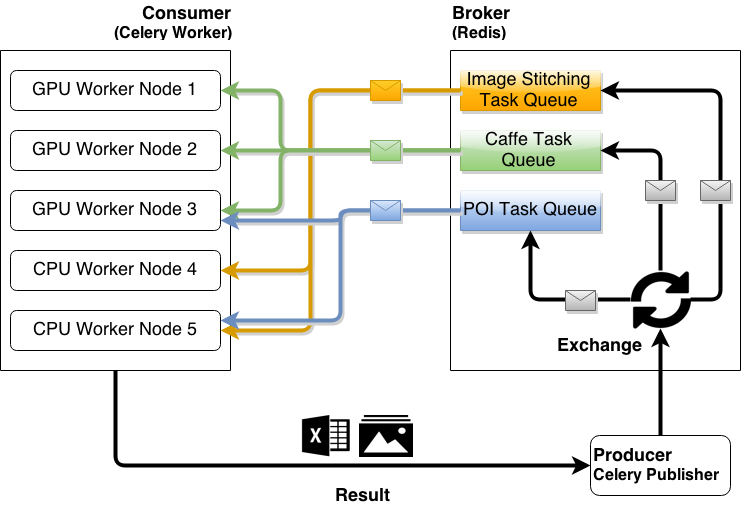}
	\caption{Celery Flow Chart.
	\label{fig:celeryflowchart}}
\end{figure*}

Celery~\cite{celery} is an asynchronous task queue based on distributed message passing. 
The execution units, called tasks, are executed concurrently on a single or 
more worker servers using their multiprocessing architecture. 
Tasks can execute asynchronously (in the background) or synchronously (wait until ready). 

CloudCV infrastructure contains heterogenous group of virtual machines that act as worker nodes, 
also called `consumers'. The master node (`producer') on receiving a job request converts the 
request into a task by serializing the input data using format such as JSON~\cite{json} and sends it to a `broker'. 
The job of the broker is to receive a task from the producer and send it to a consumer. 
Broker consists of two components: an exchange and queues. 
Based on certain bindings or rules, exchange sends each task to a particular queue. 
For instance, GPU optimized tasks (such as image classification \secref{sec:classification} 
are sent to `Classification Queue' which are then processed by worker nodes that have GPUs. 
On the other hand, image stitching tasks that utilize multiple CPUs are sent to CPU-only 
machines via `Image Stitching Queue'. 
A queue is simply a buffer that stores the messages. 

This protocol is known as AMQP Protocol~\cite{amqp} and Celery abstracts 
away details of the protocol efficiently, allowing the system to scale. 

\subsubsection{GraphLab}

GraphLab~\cite{low_uai10} 
is a high-level abstraction for distributed computing that efficiently and intuitively 
expresses data and computational dependencies with a sparse data-graph. 
Unlike other abstractions such as Map-Reduce, 
computation in GraphLab is expressed as a vertex program, which is executed in parallel on each vertex 
(potentially on different machines), while maintaining data consistency between machines and appropriate locking. 

We implemented a parallel image stitching algorithm by creating GraphLab wrappers
for the image stitching pipeline in OpenCV~\cite{opencv}, a widely used open-source computer vision library. 
The implementation is open-source and is available in the GraphLab's computer vision toolkit~\cite{graphlab_cvkit}. 

\subsection{Caffe - Deep Learning Framework}

Caffe~\cite{jia2014caffe} is a deep learning framework initially developed by the Berkeley Vision group
and now an open-source project with multiple contributors. 
In performance tests, it consistently ranks as of the fastest Convolution Neural Network (CNN) 
implementations available online. 
Caffe is widely used in academic research projects, 
industrial applications pertaining to vision, speech, \etc.  
A number of state-of-the-art models implemented in Caffe are available publicly available for download. 

CloudCV uses Caffe at the back-end to provide services such as classification, 
feature extraction and object detection. 
CloudCV also allows adding a new category to a pre-trained CNN model without retraining the entire model and is described in detail in \secref{trainaclass}.

\subsection{Front-end Platforms}

CloudCV computer vision algorithms are accessible via three front-end platforms: 
1) Web interface, 2) Python APIs, and 3) Matlab APIs. 

\begin{figure*}[t]
	\centering
	\begin{subfigure}[b]{0.49\textwidth}
		\includegraphics[width=1\columnwidth]{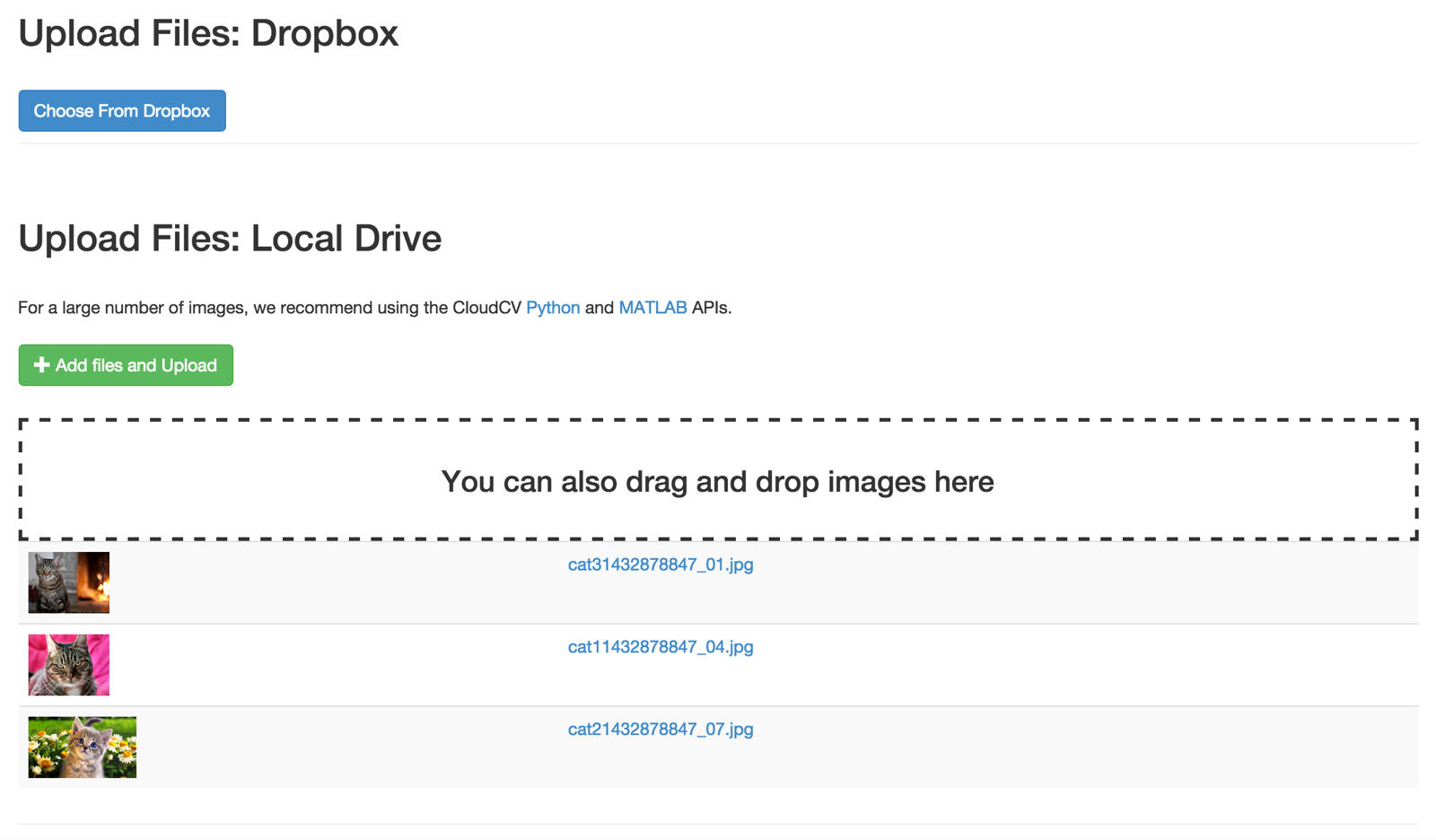}
		\subcaption{Upload interface
		\label{fig:webinterfaceupload}}
	\end{subfigure} 
	\,
	\begin{subfigure}[b]{0.49\textwidth}
		\includegraphics[width=1\columnwidth]{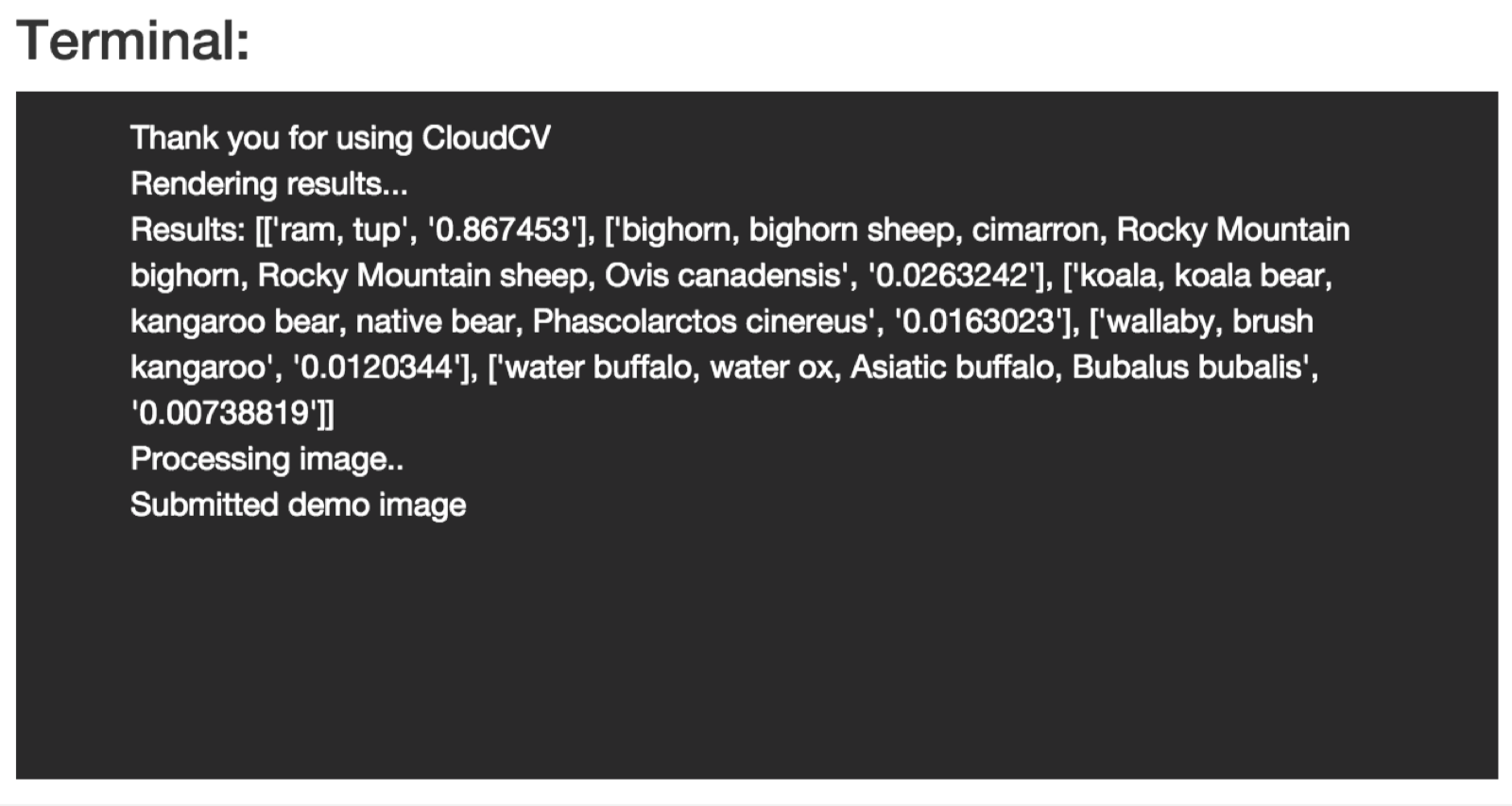}
		\subcaption{Terminal updates
		\label{fig:webinterfaceterminal}}
	\end{subfigure} 
	\,
	\begin{subfigure}[b]{1\textwidth}
		\includegraphics[width=1\columnwidth]{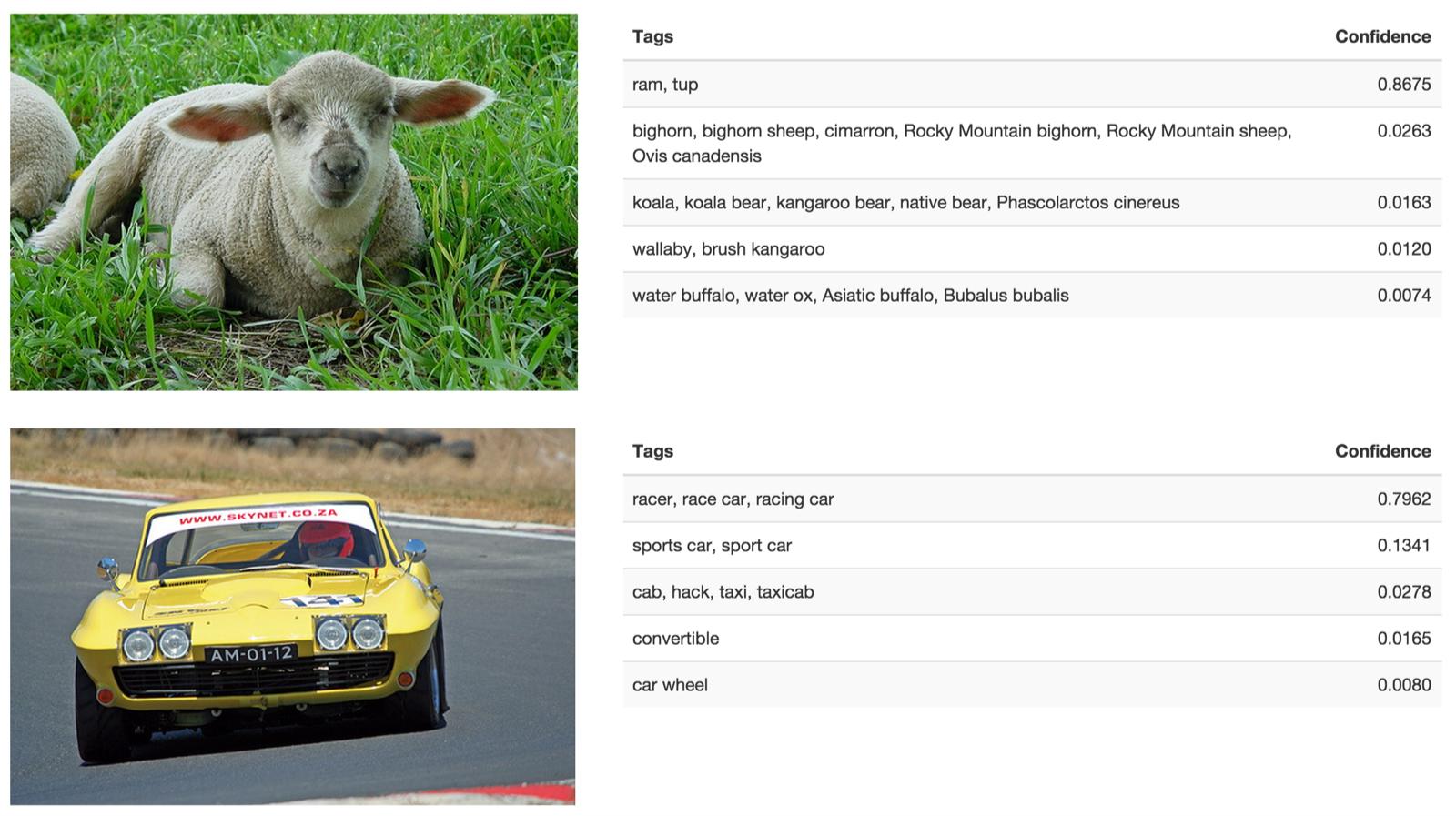}
		\subcaption{Result Visualization
		\label{fig:webinterfaceresult}}
	\end{subfigure}
	\caption{(a) shows the upload section. User can upload images either from his/her dropbox or local disk, (b) shows the  terminal which receives real-time progress updates from the server and (c) shows the visualization interface for a given job. Note in this case the task was classification and the result displays category and corresponding confidence for a given image.
	\label{fig:webinterface}}
\end{figure*}

\subsubsection{Web interface}

Modern web browsers offer tremendous capabilities in terms of accessing online content, 
multimedia \etc. 
We built a web-interface available at \url{http://cloudcv.org} 
so that users can access CloudCV from any device via any operating system 
without having to install any additional software. 

As illustrated in the screen capture in \figref{fig:webinterface}, 
users can test CloudCV services by trying them out on a few images uploaded 
through local system or upload images from third party cloud storage such as Dropbox
(shown in \figref{fig:dropbox}). 

We are also working on providing user authentication such that users can have access to 
all the trained models, training images, and job history. 
This will enable the user to seamlessly transfer transfer data across multiple data sources.  


\begin{figure*}[t]
	\centering
	\begin{subfigure}[b]{0.49\textwidth}
		\includegraphics[width=1\columnwidth]{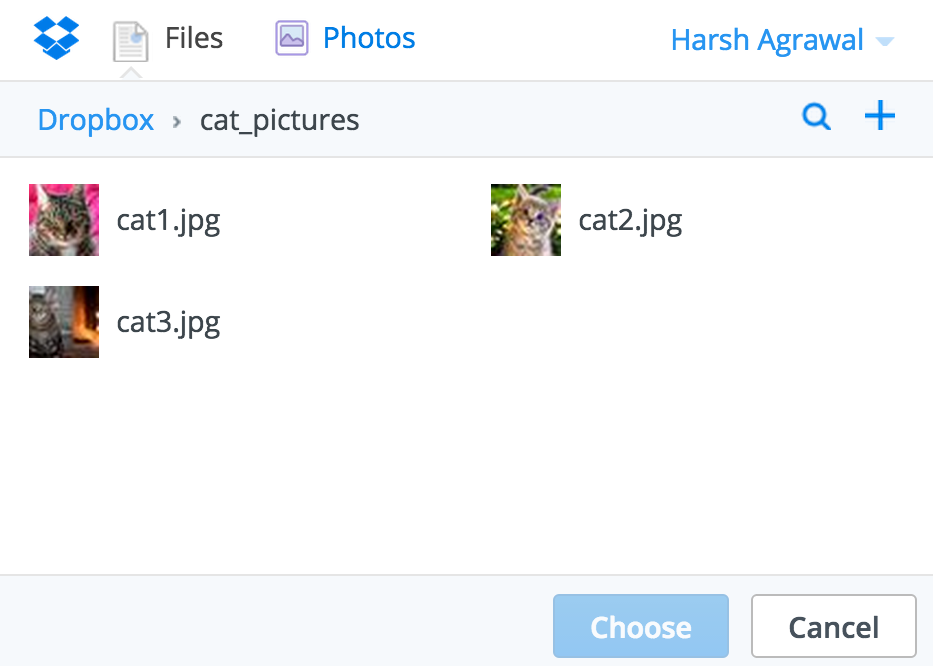}
		\subcaption{Upload interface
		\label{fig:dropboxupload}}
	\end{subfigure} 
	\,
	\begin{subfigure}[b]{0.49\textwidth}
		\includegraphics[width=1\columnwidth]{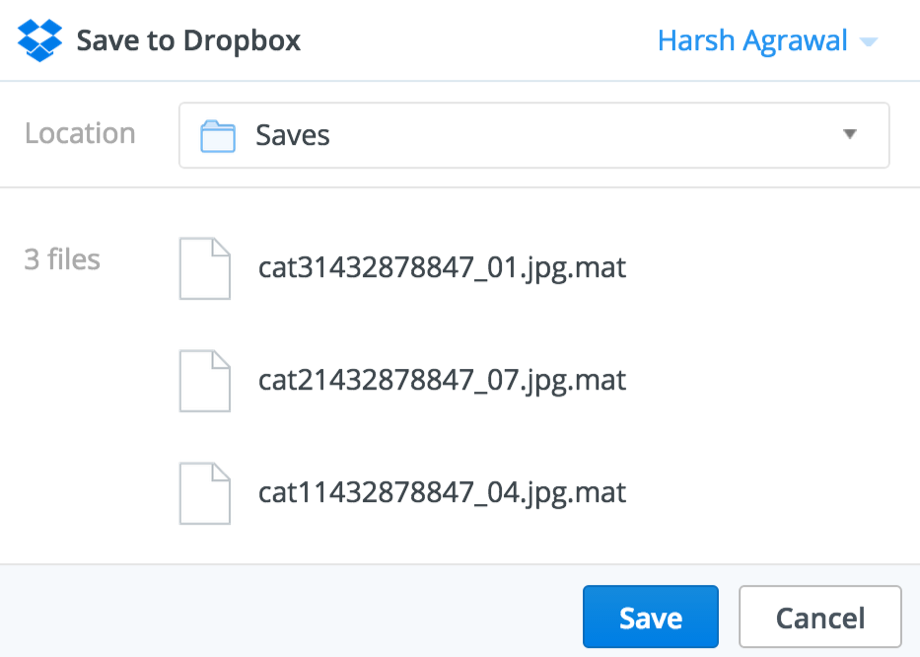}
		\subcaption{Terminal updates
		\label{fig:dropbox}}
	\end{subfigure} 
	\caption{The web-interface allows user to upload images and save features or models files inside his/her Dropbox account. (a) shows the upload interface where a user can select one or multiple images and (b) shows the save interface where the user can save all the data generated for the given job inside Dropbox. In this example, the user was trying to save features extracted in the form of Mat files inside Dropbox account. 
	\label{fig:dropbox}}
\end{figure*}

\subsubsection{Python API}

To enable building end-to-end applications on top of CloudCV, 
we make our services accessible via a Python API. 

Python has seen significant growth in terms of libraries developed for scientific 
computation because of its holistic language design. 
It also offers interactive terminal and user interface which makes data analysis, visualization and debugging easier.

\noindent \textbf{Loading necessary packages:} 
To use the CloudCV Python API, a user only needs to import the PCloudCV class. 

\begin{python}[language=python, firstnumber=1]
	from pcloudcv import PCloudCV
	import utility.job as uj
	import json
	import os
\end{python}

At this point, the pcloudcv object may be used to access the various functionalities 
provided in CloudCV. These functionalities are detailed in \secref{func}. 


\noindent \textbf{Setting the configuration path:} 
When used in the above manner, the user needs to provide 
details about the job (executable name, parameter settings, \etc) for each such API call. 
In order to reduce this burden, our API includes a \emph{configuration file} 
that stores all such necessary job information. 
The contents of this configuration file are shown below. 

\begin{code}[language=json,firstnumber=1]
{
    "exec": "classify",
    "maxim": 500,
    "config": [
        {
            "name": "ImageStitch",
            "path": "dropbox:/1/",
            "output": "/home/dexter/Pictures/test_download",
            "params": {
                "warp": "plane"
            }
        },
        {
            "name": "classify",
            "path": "local: /home/dexter/Pictures/test_download/3",
            "output": "/home/dexter/Pictures/test_download",
            "params": {
            }
        },
        {
            "name": "features",
            "path": "local: /home/dexter/Pictures/test_download/3",
            "output": "/home/dexter/Pictures/test_download",
            "params": {
                "name": "decaf",
                "verbose": "2",
            }
        }
    ]
}
\end{code}

The user may simply provide the full path to the configuration file. 
\begin{python}[language=python, firstnumber=11]
#full path of the config.json file
config_path = os.path.join(os.getcwd(), "config.json") 
dict = {"exec": "classify"}
\end{python}

\noindent \textbf{Creating PCloudCV object:} 
To run a job, the user simply needs to create a PCloudCV object. 
The constructor takes the path to the configuration file, a dictionary that contains 
optional settings for input directory, output directory, and executable, and a boolean 
parameter that tells the API whether the user wishes to login to his/her account 
using third party authorization -- Google accounts or Dropbox. 
If the boolean parameter is false, then the job request is treated as anonymous.  

\begin{python}[language=python, firstnumber=9]
p = PCloudCV(config_path, dict, True)
p.start()
\end{python}

\subsubsection{Matlab API}

\begin{figure*}[t]
	\centering
	\includegraphics[width=1\columnwidth]{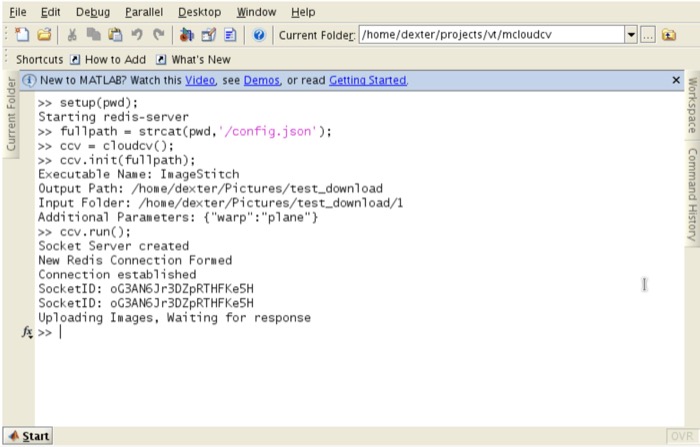}
	\caption{ 
	MATLAB API Screenshot: Users can access CloudCV services within MATLAB. These APIs run in background such that while the user is waiting for a response, the user can run other tasks and the API call is non-blocking. \label{fig:matlabScreenshot} shows the screenshort of the MATLAB API.}
\end{figure*}

MATLAB is a popular high level language and interactive environment that 
offers high-performance numerical computation, data analysis, visualization capabilities, 
and application development tools. 
MATLAB has become a popular language in academia, especially 
for computer vision researchers, because it provides easy access to thousands 
of low-level building-block functions and algorithms written by experts in addition 
to those specifically written by computer vision researchers. 
Therefore, CloudCV includes a MATLAB API, as shown in the screenshot.\figref{fig:matlabScreenshot}

\section{CloudCV Functionalities}
\label{func} 
We now describe the functionalities and algorithms currently implemented in CloudCV. 

\subsection{Classification}
\label{sec:classification}
`Image Classification' refers to predicting the class labels of objects present in an image. 
This finds myriad applications in visual computing. 
Knowing what object is visible to the camera is an immense capability in mobile applications. 
CloudCV image classification tackles this problem in the cloud. 
The classification API can be invoked to get a list of top five objects present in the image 
with the corresponding confidence scores. 

\begin{figure*}[t]
\centering
\includegraphics[width=0.9\linewidth]{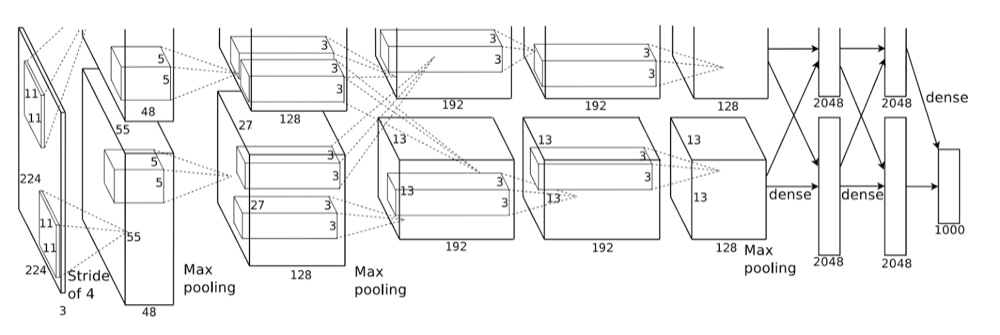}
\caption{Caffenet model architecture} 
\label{fig:caffenet}
\end{figure*}

\begin{figure*}[t]
\centering
\includegraphics[width=0.9\linewidth]{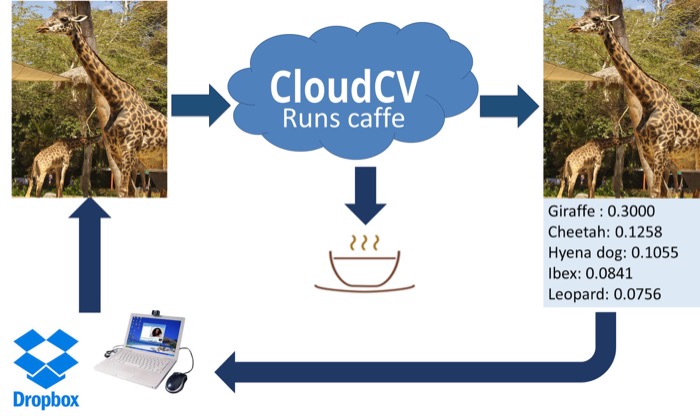}
\caption{Classification Pipeline} 
\label{fig:classification}
\end{figure*}

CloudCV classification implementation uses the `caffenet' model 
(bvlc\_referenc\-e\_caffenet in Caffe) shown in \figref{fig:caffenet} 
which is based on AlexNet~\cite{krizhevsky_nips12} architecture. 
The AlexNet architecture consists of 5 convolutional layers and 3 fully connected layers. 
The last fully connected layer (also known as fc8 layer) 
has 1000 nodes, each node corresponding to one ImageNet category. 

\subsection{Feature Extraction}

It has been shown \cite{Donahue_ICML2014, razavianASC14} that features extracted from the 
activation of a deep convolutional network trained in a fully-supervised fashion 
on an image classification task (with a fixed but large set of categories) 
can be utilized for novel generic tasks that may differ significantly from the original 
task of image classification. 
These features are popularly called DeCAF features. 
A computer vision researcher who just needs DeCAF features on his dataset, 
is currently forced to set up the entire deep learning framework, which may or may not be 
relevant to them otherwise. 
CloudCV alleviates this overhead by providing APIs that can be used to 
extract DeCAF features on the cloud and then download them as a `mat' file for further use. 

\begin{figure*}[t]
\centering
\includegraphics[width=0.9\linewidth]{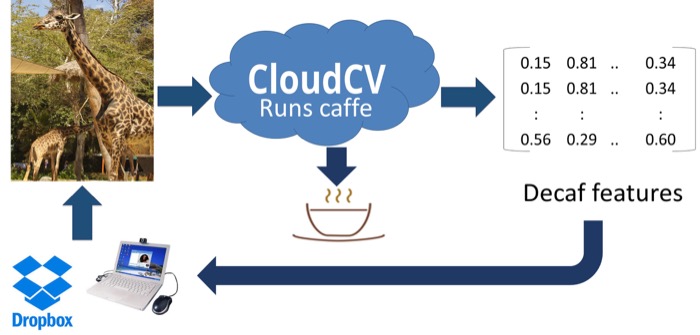}
\caption{Feature Extraction Pipeline} 
\label{fig:feature}
\end{figure*}

CloudCV feature extraction implementation uses the same architecture as \figref{fig:caffenet}. 
The DeCAF features are the activations in the second-last fully connected layer 
(also known as fc7 layer), which consists of 4096 nodes. 
The caffenet model uses the fc7 activations computed from of 10 sub-images -- 4 corner regions, 
the center region and their horizontal reflections. Therefore, 
the output is a matrix of size (10,4096). 
\subsection{Train a New Category}
\label{trainaclass}
The classification task described above is limited to a pre-defined set of 1000 ImageNet categories. 
In a number of situations, a user may need a classification model with categories other than ImageNet 
but may not have sufficient data or resources to train a new model from scratch.
CloudCV contains a `Train a New Category' capability that can be used to efficiently add new 
categories to the existing caffenet model with 1000 ImageNet categories. 

\begin{figure*}[t]
\centering
{\includegraphics[width=1\linewidth]{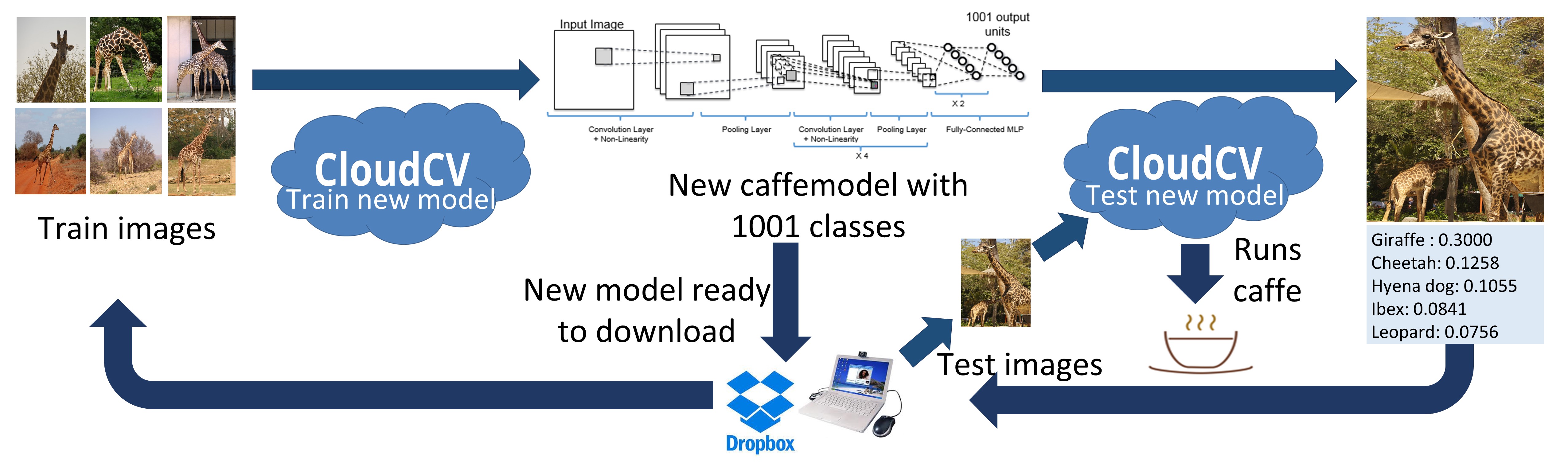}}
\caption{Train a New Category pipeline} 
\label{fig:trainaclass}
\end{figure*}

The new model is generated by appending additional nodes in the last fully connected 
layer (fc8 layer) of the existing caffenet model. 
Each new node added corresponds to a new category. 
The weights and biases for these additional nodes are computed using 
Linear Discriminant Analysis (LDA), which is equivalent to learning a 
Gaussian Naive Bayes classifier with equal covariance matrices for all categories. 
All other weights and biases are kept same as the existing caffenet model. 

The LDA weight vector ($w_k$) and bias ($b_k$) for a new category k are computed as: 
\begin{equation}
\begin{split}
w_k & = \Sigma^{-1} \mu_k \\
b_k & = \log \pi_k - \frac{1}{2} \mu^T_k \Sigma^{-1} \mu_k
\end{split}
\end{equation}
where, $\pi_k$ is the prior probability of kth category, 
$\Sigma$ is the covariance matrix of fc7 (second last fully connected layer in caffenet model) feature vectors, 
and $\mu_k$ is the mean vector of fc7 feature vectors of the given training images for the new category. 
The prior distribution is assumed to be uniform for this demo, thus the prior probability $\pi_k$ 
is just the reciprocal of number of categories. 
Notice that the covariance matrix $\Sigma$ can be computed offline using all images in the ImageNet 
training dataset, and its inverse can be cached. This is the most computationally expensive step 
in calculating the new parameters (weights and biases), but is done once offline. 
The mean vector $\mu_k$ is computed from the training images for the new category in real time, 
Thus, a new category can be added to the network instantaneously! 

We have also experimented with fine-tuning the softmax layer, and the entire network from 
this LDA initialization, however, that is useful only when significant training data is available for 
the new category.

\subsection{VIP: Finding Important People in Group Images}

\begin{figure}[t]
\begin{subfigure}[c]{0.48\columnwidth}
\includegraphics[width=0.98\textwidth]{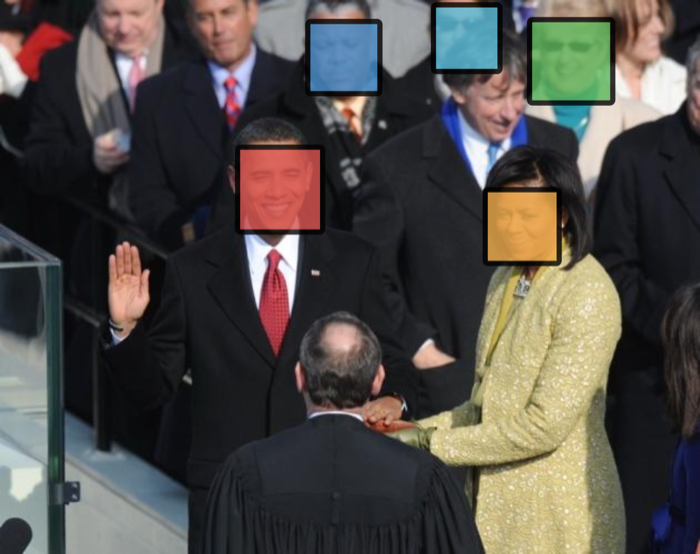}
\end{subfigure}
\begin{subfigure}[t]{0.48\columnwidth}

\includegraphics[width=0.98\textwidth]{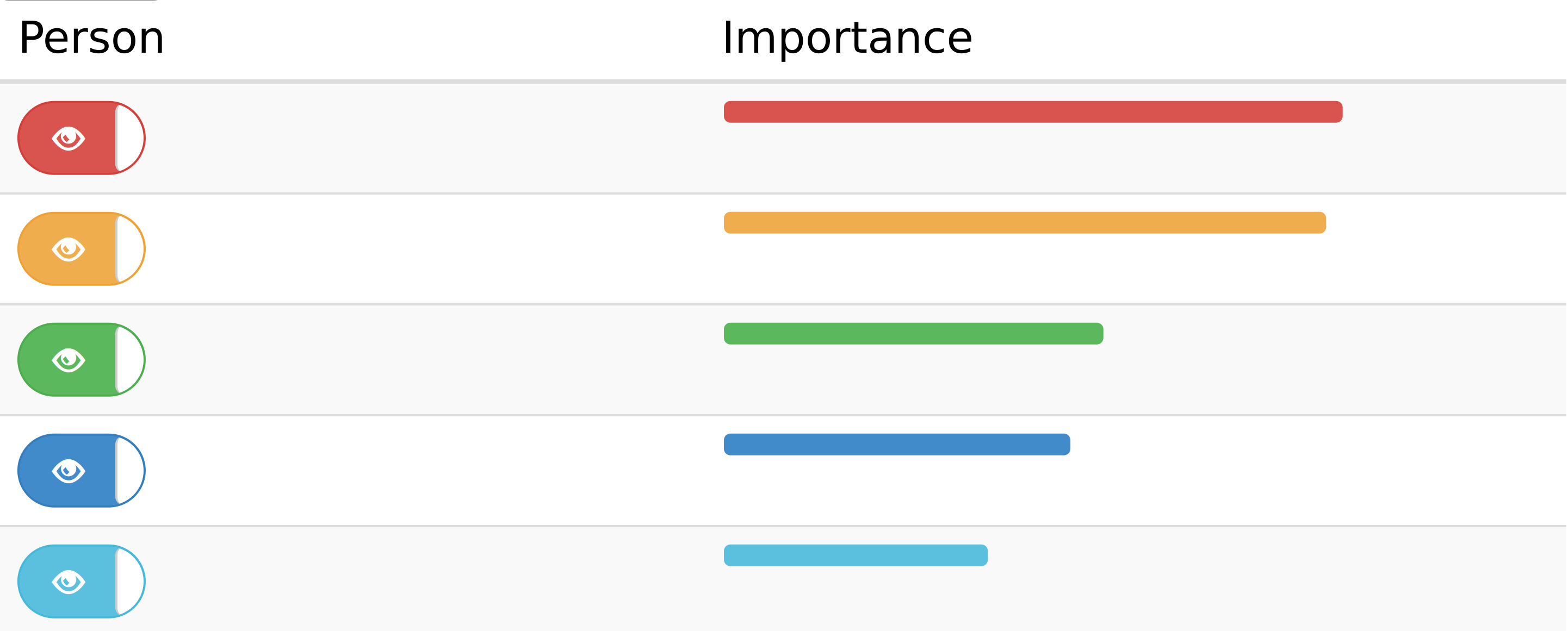}
\end{subfigure}
\caption[VIP]{VIP: Predict the importance of individuals in group photographs 
(without assuming knowledge about their identities).} 
\label{fig:poiTeaser}
\end{figure}

When multiple people are present in a photograph, there is usually a story behind the situation 
that brought them together: a concert, a wedding, 
a presidential swearing-in ceremony (\figref{fig:poiTeaser}), 
or just a gathering of a group of friends. In this story, not 
everyone plays an equal part. Some person(s) are the main character(s) and play a more 
central role.

\begin{figure*}[t]
\centering
\begin{subfigure}[t]{0.32\textwidth}
\includegraphics[width=0.9\textwidth, height=28mm]{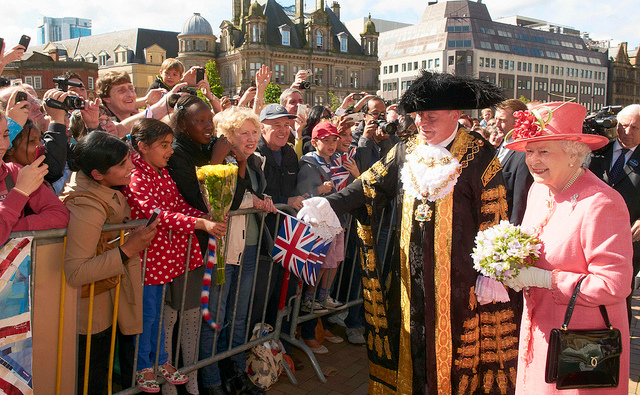}
\caption{\footnotesize Socially prominent people.}
\label{fig:poi1}
\end{subfigure}
\hfill
\begin{subfigure}[t]{0.32\textwidth}
\includegraphics[width=0.9\textwidth, height=28mm]{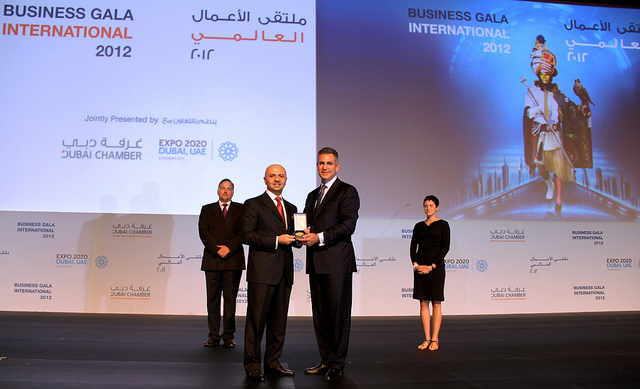}
\caption{\footnotesize Non-celebrities.}
\label{fig:poi2}
\end{subfigure}
\hfill
\begin{subfigure}[t]{0.32\textwidth}
\includegraphics[width=0.9\textwidth, height=28mm]{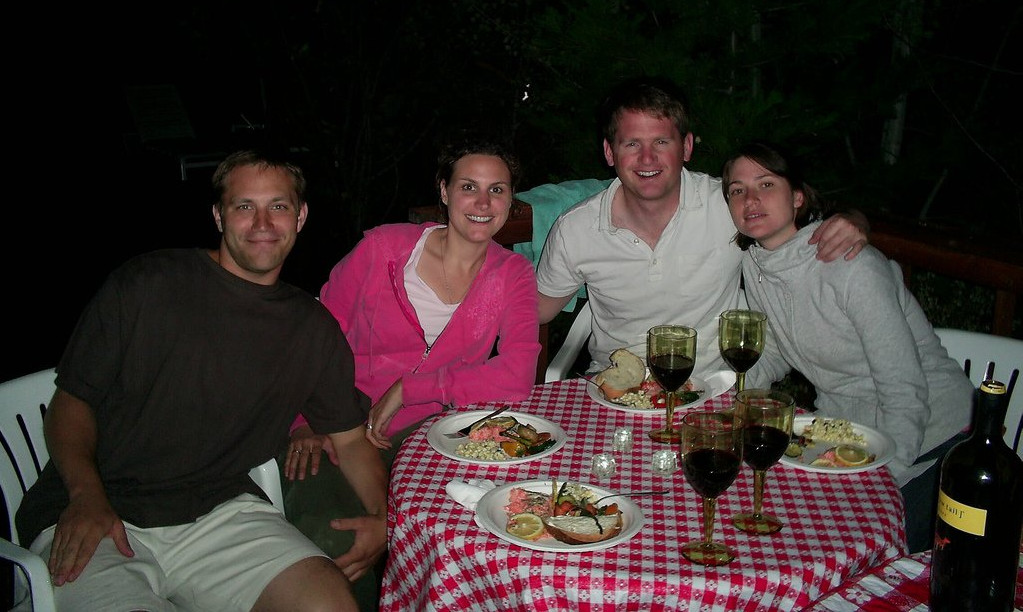}
\caption{\footnotesize Equally important people.}
\label{fig:poi3}
\end{subfigure}
\label{fig:poiExamples}

\caption[Who are the most important individuals in these pictures?]{
Who are the most important individuals in these pictures?
(a) the couple (the British Queen and the Lord Mayor);  
(b) the person giving the award and the person receiving it play the main role;
(c) everyone seems to be nearly equally important. 
Humans have a remarkable ability to understand social roles and identify important players, 
even without knowing identities of 
the people in the images. 
}

\end{figure*}

Consider the picture in \figref{fig:poi1}. Here, the important 
characters are the couple who appear to be the British Queen and the Lord Mayor. 
Notice that their identities and social status play a role in establishing their positions as the 
key characters in that image. However, it is clear that even someone unfamiliar with the oddities and eccentricities of the British Monarchy,  who simply views this as a picture of an elderly woman and a gentleman in costume  
receiving attention from a crowd, would consider 
those two to be central characters in that scene.

\figref{fig:poi2} shows an example with people who do not appear to be celebrities. 
We can see that two people in foreground are clearly the focus of attention, and 
two others in the background are not. 
\figref{fig:poi3} shows a common group photograph, where everyone is nearly equally important.
It is clear that even without recognizing the identities of people, we as humans 
have a remarkable ability to understand social roles and identify important players. 

\noindent \textbf{Goal.}
The goal of CloudCV VIP is to \emph{automatically predict the importance of individuals in group photographs}. 
In order to keep our approach general and applicable to any input image, 
we focus purely on visual cues available in the image, and do not assume identification of 
the individuals. Thus, we do not use social prominence cues. For example, in \figref{fig:poi1}, 
we want an algorithm that identifies the elderly woman and the gentleman as the two most important 
people that image without utilizing the knowledge that the elderly woman is the British Queen.

A number of applications can benefit from knowing the importance of people. 
Algorithms for im2text (generating sentences that describe an image) can be made more human-like if they 
describe only the important people in the image and ignore unimportant ones. 
Photo cropping algorithms can do ``smart-cropping''  of images of people by keeping 
only the important people. Social networking sites and image search applications can benefit from improving the ranking of photos where 
the queried person is important. Intelligent social robots can benefit from identifying important people in any scenario.

\noindent \textbf{Who is Important?}
In defining importance, we can consider the perspective of three parties (which may disagree):

\begin{itemize}
\item \textbf{the photographer}, who presumably intended to capture some subset of people, 
and perhaps had no choice but to capture others; 
\item \textbf{the subjects}, who presumably arranged themselves following social inter-personal rules;  and

\item \textbf{neutral third-party human observers}, who may be unfamiliar with the subjects of the photo 
and the photographer's intent, but may still agree on the (relative) importance of people. 

\end{itemize}

\begin{figure}[t]
\includegraphics[width=0.98\columnwidth]{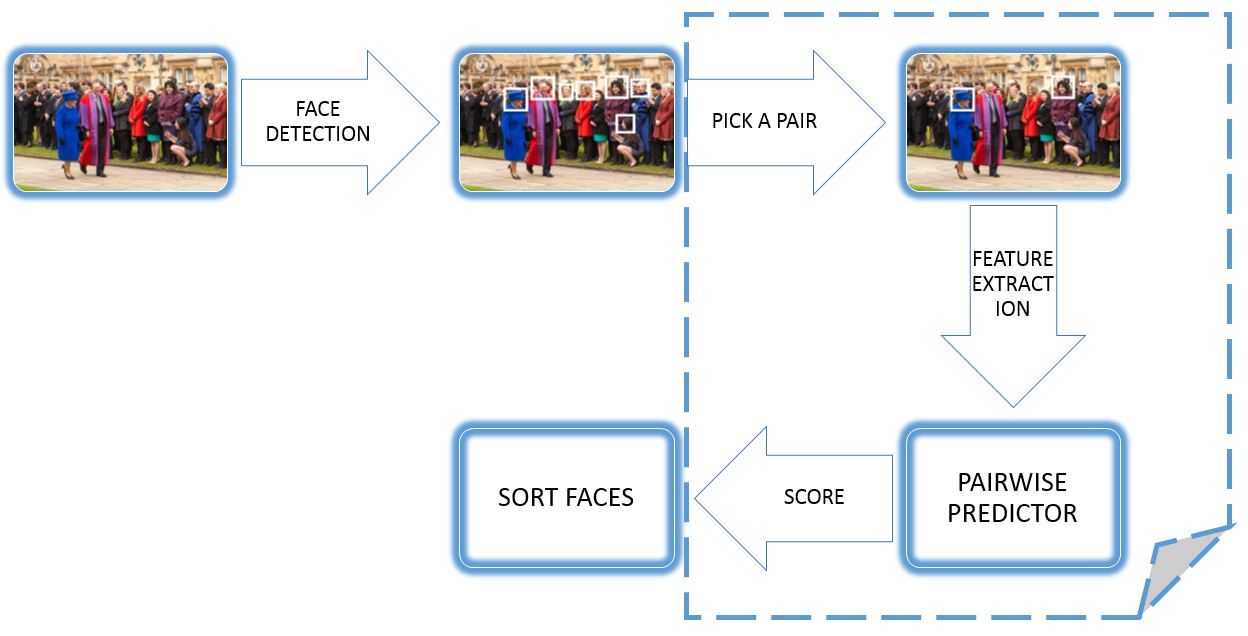}
\caption[VIP Pipeline]{VIP Pipeline.} 
\label{fig:poiArch}
\end{figure}

Navigating this landscape of perspectives involves many complex social relationships: the social status of each person in the image (an award winner, a speaker, the President), and 
the social biases of the photographer and the viewer (\eg, gender or racial biases); many of these can not be easily mined from the photo itself. 
At its core, the question itself is subjective: if the British Queen ``photo-bombs'' while you are taking a picture of your friend, 
is she still the most important person in that photo? 

In CloudCV VIP, to establish a quantitative protocol, we rely on the wisdom of 
the crowd to estimate the ``ground-truth'' importance of a person in an image. 
Our relative importance models are trained using real-valued importance scores 
obtained using Amazon Mechanical Turk.

\noindent \textbf{Pipeline.} 
The basic flow of CloudCV VIP is shown in \figref{fig:poiArch}. 
First, face detection is performed using third-party face detectors. 
In our published work~\cite{vip}, we used Sky Biometry's API~\cite{skybiometry} for face detection. 
CloudCV VIP uses OpenCV~\cite{opencv_library} to avoid network latency. 
For every pair of detected faces, features are extracted that describe the relative configuration of these faces. 
These features are fed to our pre-trained regressors to derive a relative importance score for this pair. 
Finally, the faces are sorted in descending order of importance. 
The models and features are described in detail in Mathialagan~\etal~\cite{vip}. 
In order to be fast during test time, CloudCV VIP does not use DPM based pose features.

\subsection{Gigapixel Image Stitching}

The goal of \emph{Image Stitching} is to create a composite panorama from a collection of images. 
The standard pipeline~\cite{brown_ijcv07} for Image Stitching,
consists of four main steps: 

\begin{packed_enum}

\item{Feature Extraction:} distinctive points (or keypoints) are identified in each image 
and a feature descriptor~\cite{sift,surf} is computed for each keypoint. 

\item{Image/Feature Matching:} features are matched between pairs of images to 
estimate relative camera transformations. 

\item{Global Refinement:} of camera transformation parameters across all images.

\item{Seam Blending:} seams are estimated between pairs of images and blending is performed. 
\end{packed_enum}

\begin{figure*}[h]
\centering
\begin{subfigure}[b]{1\textwidth}
	{\includegraphics[width=1\linewidth]{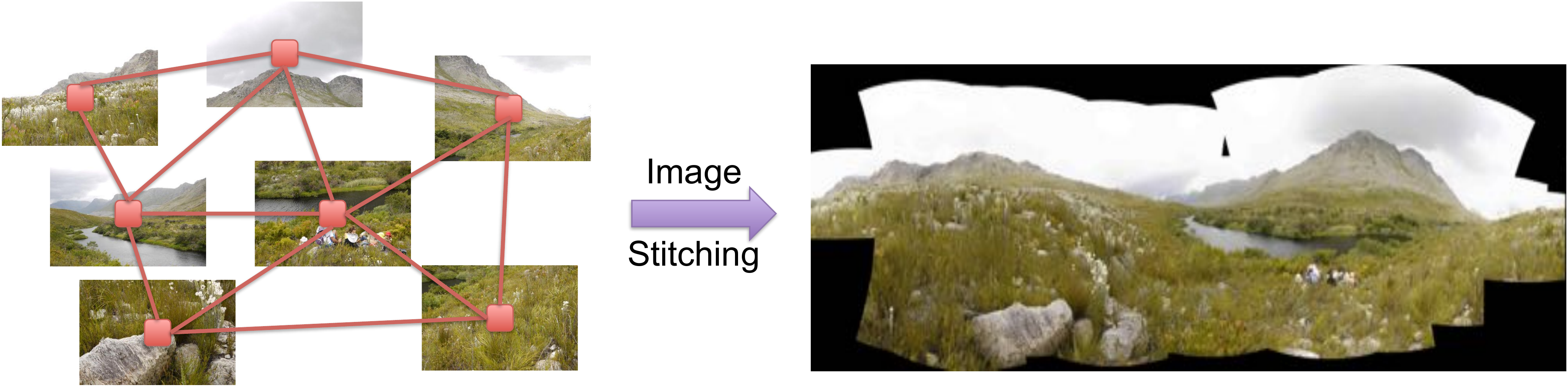}}\\
	\subcaption{Data-Graph and Panorama.
	\label{fig:stitchpana}}
\end{subfigure}
\begin{subfigure}[b]{1\textwidth}
	{\includegraphics[width=1\linewidth]{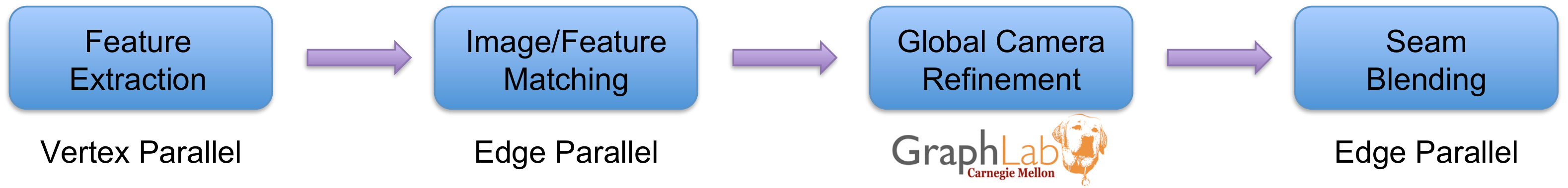}}\\
	\subcaption{Stitching Pipeline.
	\label{fig:stitchpanb}}
\end{subfigure}
\caption{ Gigapixel Image Stitching
\label{fig:stitchpan}}
\end{figure*}

Consider a data graph $G = (V,E)$ 
where each vertex corresponds to a image and two vertices are connected by 
an edge if the two images overlap in content, \ie capture a part of the scene from two viewpoints. 
In the context of this graph, different steps of the stitching pipeline have vastly different levels of 
parallelism. Step 1 (feature extraction) is \emph{vertex-parallel} since 
features extraction at each vertex/image may be run completely independently. Step 2 
(image/feature matching) and step 4 (seam blending) are \emph{edge-parallel} since these computations
may be performed completely independently at each edge in this graph. Together these steps are 
\emph{data-parallel}, where parallelism is achieved simply by splitting the data onto different machines
with no need for coordination. 

Step 3 (global refinement) is the most sophisticated step since it not embarrassingly parallel. 
This global refinement of camera parameters, 
involves minimizing 
a nonlinear error function (called re-projection error) that necessarily depends on all images~\cite{triggs_va99}, 
and ultimately slows the entire pipeline down. 
 
We formulate this optimization as a ``message-passing'' operation on the data graph -- 
each vertex simply gathers some quantities from its neighbors and makes local updates
to its camera parameters. Thus, different image can be processed on different machines 
as long as they communicate their camera parameters to their neighbors. 

Thus, while this pipeline may not be data-parallel, it is \emph{graph-parallel} show in ~\figref{fig:stitchpanb}, 
meaning that data and computational dependencies are captured by a sparse undirected graph 
and all computation can be written as \emph{vertex-programs}. It is clear that thinking 
about visual sensing algorithms as vertex programs is a powerful abstraction. 

\begin{figure*}[h]
\centering
\begin{subfigure}[b]{0.48\textwidth}
	{\includegraphics[width=1\columnwidth]{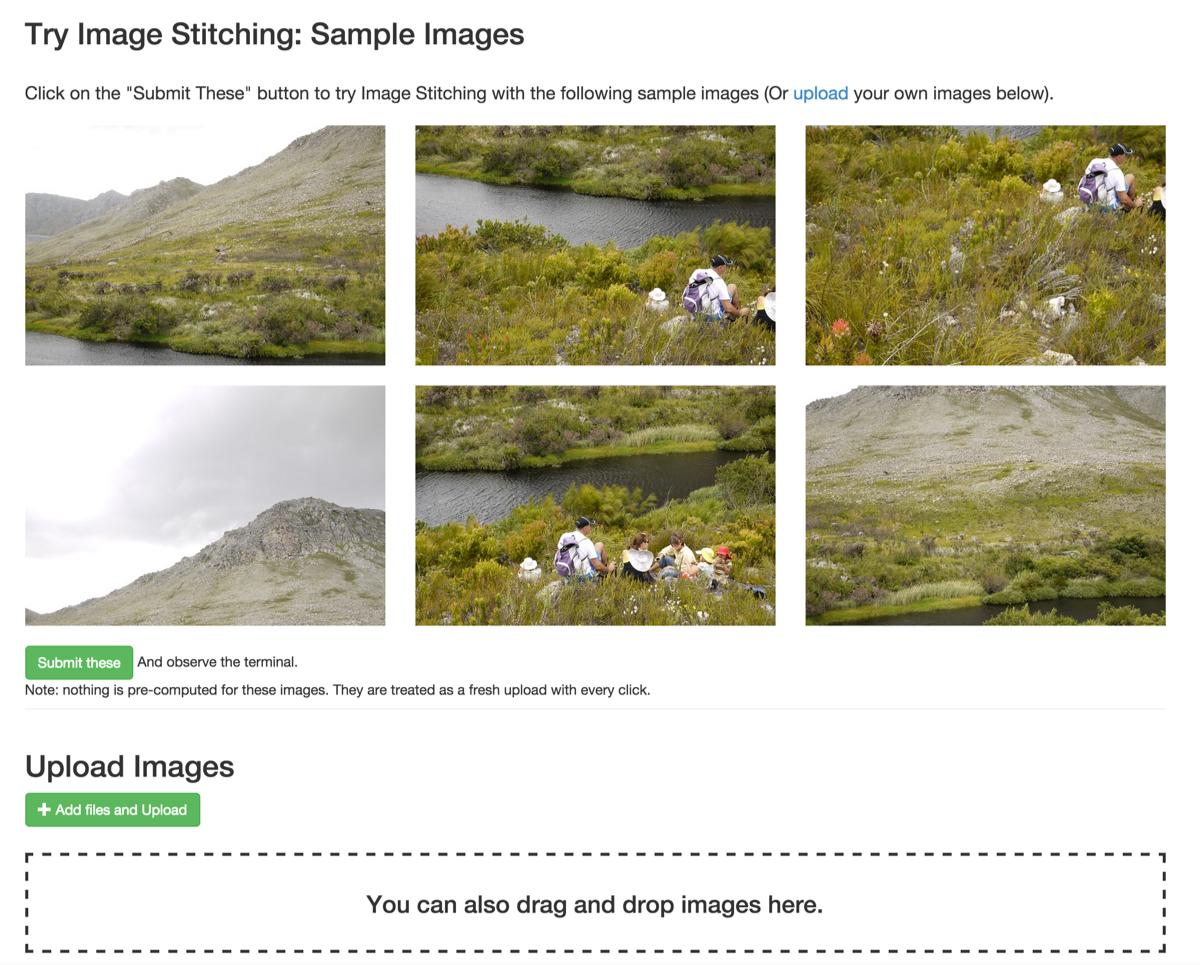}}\\
	\subcaption{Sample images and upload interface.
	\label{fig:stitchinterface}}
\end{subfigure}
\begin{subfigure}[b]{0.48\textwidth}
	{\includegraphics[width=1\columnwidth]{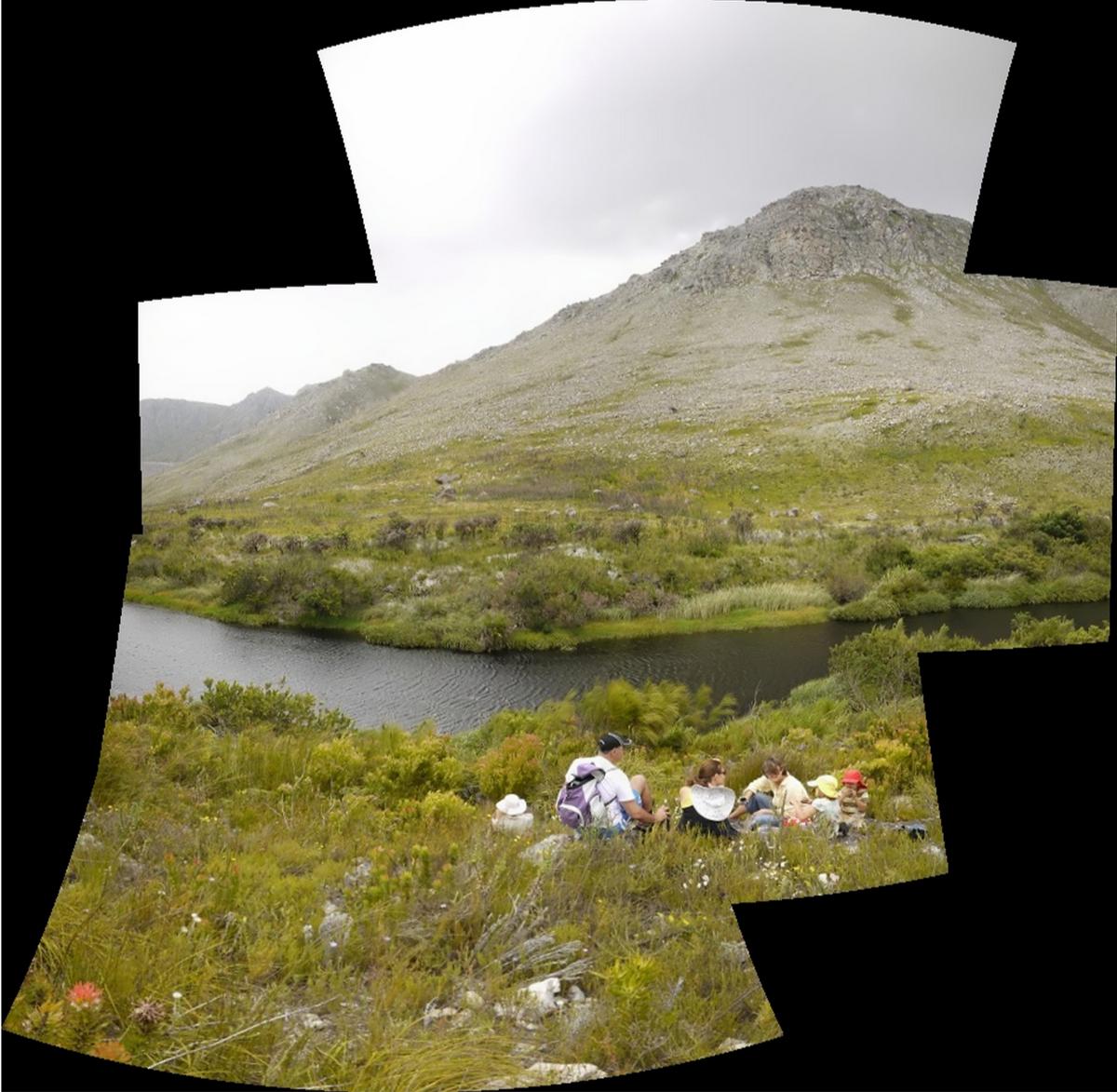}}\\
	\subcaption{Result for the sample images.
	\label{fig:stitchresult}}
\end{subfigure}
\caption{Image stitching web interface. 
\label{fig:stitchscreenshot}}
\end{figure*}

The CloudCV image stitching functionality can be accessed through the web-interface, 
a screenshot of which is shown in \figref{fig:stitchscreenshot}

\section{Future Work}
\label{sec:future_work}
\subsection{Deep Learning GPU Training System - DIGITS}
Convolutional Neural Networks (CNNs) have attracted significant interest from 
researchers in industry and academia, which has resulted in multiple 
software platforms for configuring and training CNNs. 
Most notable software platforms are Caffe, Theano, Torch7 and CUDA-Convnet2.

Recently NVIDIA\textsuperscript{TM} released DIGITS \cite{DIGITS},  an interactive Deep Learning GPU Training System, 
which provides access to the rich set of functionalities provided by Caffe \cite{jia2014caffe}
through an intuitive browser-based graphical user interface. 
DIGITS complements Caffe functionality by introducing an extensive set of visualizations 
commonly used by data scientists and researchers. Moreover since DIGITS runs as a web service, this facilitates seamless collaboration between large teams enabling data, trained CNNs model and results sharing.

\subsubsection{DIGITS Overview}
A typical work-flow of DIGITS can be summarized as follows:

\begin{itemize}
\item \textbf{Data Source}: First the user has to upload the database to be use for training and validation to DIGITS. Currently the database of image have to be stored on the same local machine hosting DIGITS.

\item \textbf{Network Architecture:} Two options are supported for network architecture:
\begin{itemize}
\item to use a standard network architecture such as AlexNet \cite{krizhevsky2012imagenet}
\item create a customized CNN where user can define each layer and associated parameters. 
\end{itemize}
DIGITS also provides a tool to visualize the CNN architecture for visual inspection of the network topology.

\item \textbf{Training and Visualization:} Users can train the newly defined network 
and track its progress real-time. The learning rate and accuracy of the model can be 
seen from real-time graph visualizations as training progresses. 
The user can abort training anytime if he suspects there is bug in the network configuration. 
Moreover similar to Caffe, DIGITS save multiple snapshots of the CNN as training progress 
giving the user the option to use specific snapshots to generate feature extraction. 
 \end{itemize}

 \subsubsection{Integrating DIGITS with CloudCV}

CloudCV classification functionality uses Caffe as the software platform to train and configure CNN. One of the main objectives of CloudCV is to enable non-experts the ability to use computer vision algorithms as a service through a rich set of API. In future work we will integrate CloudCV functionalities with DIGITS intuitive graphical user interface to provide an end-to-end system that can train a model, classify or extract features from these trained model, visualize results, \etc. Future work include:

\begin{itemize}
\item Integrating the DIGITS codebase with CloudCV.
\item Currently DIGITS only support creating training data on the host machine. We plan to further extend data sources to include cloud storage such as Dropbox and Amazon S3.
\item Integerating DIGITS with Scikit-learn \cite{scikit-learn} - a rich machine learning python library - to train different classifiers (Support Vector Machine, \etc) on features extracted from intermediate layers of a CNN. Users will be able to tune the parameters of the classifier and see improvement in real-time. 
\item Extending Visualizations provided by DIGITS to include side by side plots to visualize performance of different classifiers on a specific data set.
\item Supporting for a user workspace, where each registered user can have a private workspace. User data like job history, previous datasets, pre-trained models, previous outputs will be saved.
\end{itemize}


\vspace{10pt}
\noindent 
\textbf{Acknowledgements.} 
This work was partially supported by 
the Virginia Tech ICTAS JFC Award, 
and the National Science Foundation CAREER award IIS-1350553.  
The views and conclusions contained herein are those of the authors and should not be 
interpreted as necessarily representing the official policies or endorsements, either expressed or implied, 
of the U.S. Government or any sponsor. 

{\small
\bibliographystyle{ieee}
\bibliography{cloudcv}
}

\end{document}